\documentclass[11pt]{elsarticle}

\usepackage{times}
\usepackage{latexsym}
\usepackage{graphicx}
\usepackage{subcaption}

\usepackage[T1]{fontenc}

\usepackage[utf8]{inputenc}

\usepackage{microtype}

\usepackage{array}
\usepackage{pifont}
\usepackage{tabularx}
\usepackage{adjustbox}
\usepackage{multirow}
\usepackage{enumitem}
\usepackage{xspace}
\usepackage{tcolorbox}
\usepackage{booktabs,amsfonts,dcolumn}
\usepackage{hyperref}
\usepackage{url}
\usepackage{amsmath,amsthm,amsfonts,amssymb,bm,stmaryrd,bbm}
\usepackage[noorphans,vskip=0.75ex,leftmargin=2ex]{quoting}
\usepackage{enumitem}

\let\oldparagraph\paragraph
\renewcommand{\paragraph}[1]{\vspace{1em}\oldparagraph{#1}}

\hypersetup{
pdftitle={Bespoke Large Language Models for Digital Triage Assistance in Mental Health Care},
pdfauthor={Niall Taylor},
}

\title{Bespoke Large Language Models for Digital Triage Assistance in Mental Health Care}

\date{March 2024}

\begin{document}

\begin{frontmatter}

\author[inst1]{Niall Taylor}
\author[inst1]{Andrey Kormilitzin}
\author[inst1]{Isabelle Lorge}
\author[inst1]{Alejo Nevado-Holgado}
\author[inst1,inst2,inst3]{Andrea Cipriani}
\author[inst4,inst5,inst6]{Dan W. Joyce}

\affiliation[inst1]{organization={Department of Psychiatry, University of Oxford},
            city={Oxford},
            country={United Kingdom}}
\affiliation[inst2]{organization={Oxford Precision Psychiatry Lab, NIHR Oxford Health Biomedical research Centre},
            city={Oxford},
            country={United Kingdom}}   
\affiliation[inst3]{organization={Oxford Health NHS Foundation Trust, Warneford Hospital},
            city={Oxford},
            country={United Kingdom}}   
            
\affiliation[inst4]{organization={Department of Primary Care and Mental Health, University of Liverpool},
            city={Liverpool},
            country={United Kingdom}}
\affiliation[inst5]{organization={Civic Health Innovation Labs, University of Liverpool},
            city={Liverpool},
            country={United Kingdom}}
\affiliation[inst6]{organization={Mental health Research for Innovation Centre (M-RIC), Mersey Care NHS Foundation Trust},
            city={Prescot, Merseyside},
            country={United Kingdom}}

\begin{abstract}
Contemporary large language models (LLMs) may have utility for processing unstructured, narrative free-text clinical data contained in electronic health records (EHRs) -- a particularly important use-case for mental health where a majority of routinely-collected patient data lacks structured, machine-readable content.  

A significant problem for the the United Kingdom's National Health Service (NHS) are the long waiting lists for specialist mental healthcare.  According to NHS data \cite{NHS-dig-2023}, in each month of 2023, there were between 370,000 and 470,000 individual new referrals into secondary  mental healthcare services.  Referrals must be triaged by clinicians, using clinical information contained in the patient's EHR to arrive at a decision about the most appropriate mental healthcare team to assess and potentially treat these patients.  

The ability to efficiently recommend a relevant team by ingesting potentially voluminous clinical notes could help services both reduce referral waiting times and with the right technology, improve the evidence available to justify triage decisions.  

We present and evaluate three different approaches for LLM-based, end-to-end ingestion of variable-length clinical EHR data to assist clinicians when triaging referrals.  Our model is able to deliver triage recommendations consistent with existing clinical practices and it's architecture was implemented on a single GPU, making it practical for implementation in resource-limited NHS environments where private implementations of LLM technology will be necessary to ensure confidential clinical data is appropriately controlled and governed. 
\end{abstract}

\end{frontmatter}

\section{Introduction}
Interest in large language models (LLMs) has grown substantially, including for medical applications \cite{singhal2023large}.  At-scale access to LLMs such as ChatGPT\cite{openai_gpt-4_2023} and Claude 3\cite{noauthor_claude_nodate} are offered using a machine-learning-as-a-service (MLaaS) model \cite{ribeiroMLaaSMachineLearning2015} and currently, there is uncertainty and lack of clear data governance processes, policies and ethical considerations for using LLMs in this way \cite{bleaseChatGPTMentalHealthcare2023,wangEthicalConsiderationsUsing2023}. 

Consequently, as it stands now in healthcare settings, implementing a bespoke LLM-based solution will require domain adaptation on imported, open LLMs; for example, fine-tuning {RoBERTa} for a specific use-case.  In this paper, we consider assisting clinicians make decisions about ``triaging'' patients to an appropriate specialist mental healthcare (MH) team by ingesting the patient's electronic health record (EHR) data.  Mental healthcare makes extensive use of narrative recording of clinical data as unstructured ``free-text'' making LLMs a particularly useful technology for assisting clinicians in parsing and making use of volumes of textual information.   

\subsection{Clinical Context}
In the United Kingdom, almost all healthcare is operated and provided by a publicly funded, single-payer healthcare system -- the National Health Service (NHS).  The NHS is organised such that for almost all healthcare needs, a patient consults a general practitioner (GP, sometimes called a primary care or family practice physician) and a decision is made to refer the patient to a specialist service (secondary care) as required.  In England, mental healthcare is provided by the NHS using this model and is similarly stratified into primary (led by general practice), secondary (specialist community and hospital care) and tertiary services (e.g. secure forensic services).  

\begin{figure}[htp]
    \centering
    \includegraphics[scale=0.6,trim={0 17cm 0 0}]{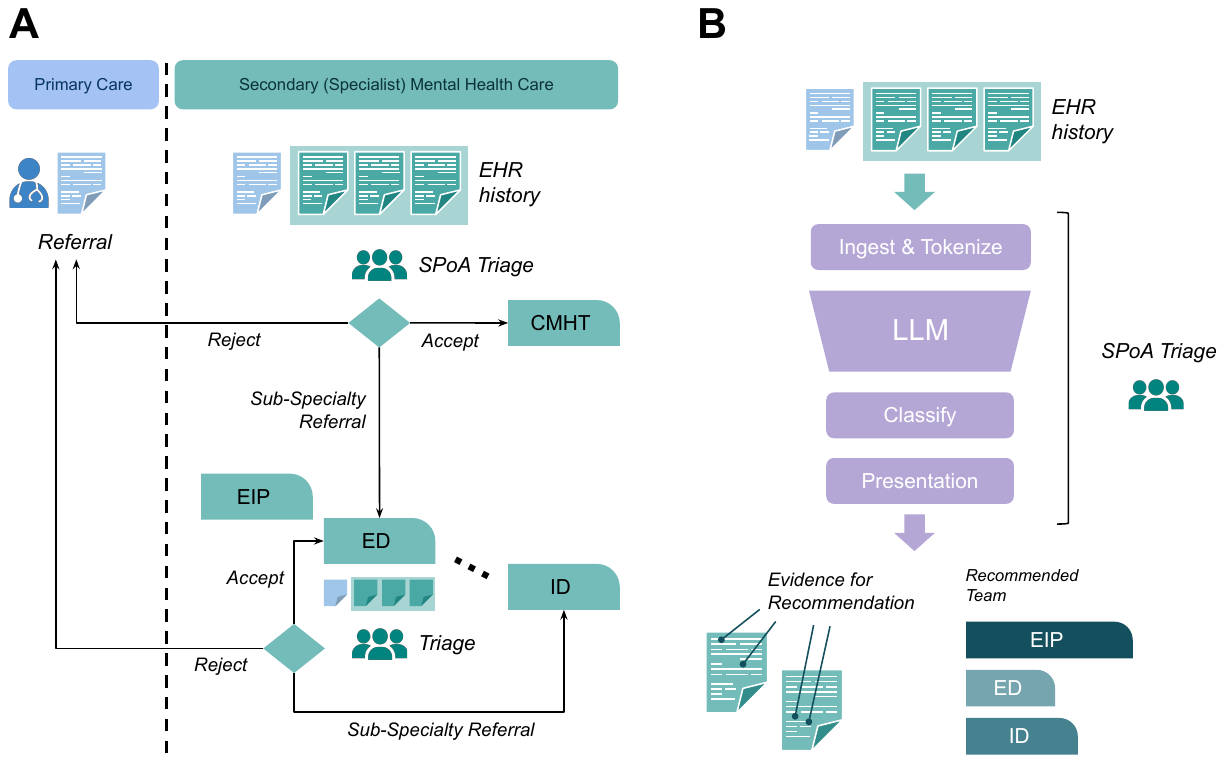}
    \caption{\textbf{A}: Schematic description of the triage process using referral and historical EHR documents highlighting the accept/reject and sub-specialty referral and re-triage process;  {SPoA} = single-point-of-access refers to the common location that receives all referrals and where a number of clinicians will perform triage \textbf{B}: End-to-end ingestion of the same clinical data for assisted triage.  Example sub-specialty teams: EIP = early intervention for psychosis, ED = eating disorders, ID = intellectual disability}
    \label{fig:triage-description}
\end{figure}

A majority of people (96\%) requiring specialist (secondary) mental healthcare are referred to -- and treated by -- community mental health teams (CMHTs) \cite{NHS-dig-2020}.  Referral to secondary care is usually via a written referral by a GP that contains a narrative of the patient's difficulties, symptoms and any relevant risks (for example, self-harm, suicide or risk posed to other people).  The NHS maintains monthly digital audits of activity in mental health services \cite{NHS-dig-2023} that show in each month of 2023, there were between 370,000 and 470,000 individual new referrals into these mental healthcare services.  Of these referrals, some will be referrals for the same patient to different clinical teams and some will reflect new referrals for patients already known to the same secondary care provider.   In most secondary care services, several CMHTs will offer a single point-of-access and triage function (see Figure \ref{fig:triage-description}), often organised to provide care to a specific geographical region.  CMHTs treat the whole spectrum of mental illness but in some circumstances, there are additional sub-specialist teams established to provide care tailored to specific conditions, for example, eating disorders or first-episode psychosis.  A complicating factor is that depending on the referring professional (e.g. a GP, a local hospital emergency department or social care professional) a patient \textit{may} be referred to a CMHT (for triage, possibly assessment and treatment) or directly to a sub-specialty team (for example, if the referrer is concerned about a specific condition such as an eating disorder, or a first episode of a psychotic disorder, they might choose to refer directly to the relevant sub-specialty team if the local secondary provider offers such a service).  A consequence of this is that patients' referrals can often be ``bounced'' between different teams; for example, a GP sees a patient who they suspect is experiencing psychotic symptoms (e.g. delusions) and refers directly to the secondary care first-episode psychosis team.  The first-episode team disagree that the referred patient is experiencing a psychotic episode, so they will forward the referral to a CMHT for further triage (see \ref{appendix:referral-bounce} for further detail).

Whenever a secondary care team (a CMHT or a sub-specialty team) receives a referral document it is generally summarised or entered onto the patient's EHR. The clinician(s) triaging the referral will also make use of any available historical information that will be contained in the patient's EHR.  This historical EHR data will exist if the patient has had previous treatment episodes, assessments or contact with the same secondary care system and will describe previous diagnoses and treatments including any hospital admissions.  Using the referral documentation as well as any available historical EHR data, the triage decisions are generally to a) ``accept'' the patient to the CMHT and proceed with a clinical assessment b) ``reject'' the referral e.g. because there is not enough information to make a decision or the clinical information suggests no role for secondary care services or c) to ``route'' the referral to a more appropriate sub-specialty team e.g. if a patient presents with psychosis or an eating disorder and there exists a sub-specialty team available for those specific conditions.  

This process is time-consuming, prone to subjective interpretation and often repeated for the same patient in different teams.  Triage requires clinical expertise and knowledge of local service arrangements and decisions are made before a patient is seen for assessment (i.e. using only recorded clinical data).  Referral and triage processes have also been criticised for a lack of transparency (to both patients and referrers), being capricious (with CMHTs using referral criteria and thresholds inconsistently) and introducing frustration for both patients and referrers from `referral bouncing' \cite{chew2007qualitative} where no team accepts the patient for further assessment, instead arguing that the patient's difficulties are more relevant to another team's remit.  This leads to so-called ``hidden waiting lists'' where patients awaiting the outcome of triage processes experience deterioration in their mental health leading and are left with no option but to seek help from emergency services, often in acute crisis \cite{HiddenWaitsForce}.

Assisted triage -- where we deploy AI in the service of improving this process -- could help by improving the efficiency of extracting and making visible the relevant clinical data and assisting in allocating and justifying triage decision making, i.e. why a given patient is suitable (or not) for any given sub-specialty team.  We stress that we are not proposing to automate triage by ingesting clinical EHR data, rather, augmenting and assisting clinicians in robustly completing the triage task.   

\section{Challenges with Unstructured EHR Data}

\subsection{Representation Learning for Locating Triage Signal in EHRs}
Due to the often noisy, idiosyncratic style of writing found in routinely-collected clinical text, clinical NLP and applications have been heavily reliant on information extraction, heuristics, and named entity recognition (NER) to locate salient information.  These have proven powerful for a number of applications such as e.g. NER for medications \cite{kormilitzin_med7_2021} and medical concept annotation \cite{kraljevic_multi-domain_2021} for delivering structured data from the original raw text. 

In contrast, triaging clinicians are able to extract relevant ``signal'' from sequences of narrative clinical notes contained in the EHR and it is reasonable to postulate that supervised representation (feature) learning  might be able to capture similar information that can be used in assisting triage tasks.  Recent transformer-based LLMs can process and represent large bodies of text capturing the semantics and pragmatics of natural language.  Importantly, we  investigate the ability to represent unprocessed, narrative clinical notes using LLMs in an end-to-end fashion where token sequences are processed to deliver feature representations (embeddings) to serve a downstream task, as opposed to information extraction.  Our approach uses LLMs to generate embeddings of patients' referral information and histories (from their EHR) in order to determine which triage team they best align with. 

To learn representations for assisting triage, a LLM must gracefully handle variable-length sequences of tokenised input, directly extracted from a patient's EHR.  In the EHR, clinical information is recorded in a time-stamped sequence of documents (notes) containing narrative and largely unstructured ``free-text''.  Each document or note can be of variable word length.  For example, a short administrative note (acknowledging the receipt of a referral, or recording a brief telephone contact with a patient) may be tens of words, whereas a clinical assessment can be thousands (illustrative descriptive statistics for our dataset are shown in \ref{appendix:referral-bounce}).


In addition, any given patient may have had one or more historical referrals or episodes of care with one or more different teams (see top panel, Figure \ref{fig:three-methods}) and these episodes will be accompanied by further EHR notes (between these referral dates) that describe the episode of care from referral to the point of discharge from services. 


\subsection{Managing Variable Token Sequence Lengths}
A common issue when using transformer-based LLMs for long sequences is the finite context window when using full self-attention \cite{longformer2020}.  The time- and memory-complexity of transformers' self-attention mechanism is a quadratic function of the length of the input sequence (i.e. the number of word tokens). In practice, this means models such as RoBERTa \cite{liu_roberta_2019, devlin-etal-2019-bert} were trained with a maximum sequence length of just 512 word tokens. 

In this work, we use three different approaches each with different limitations and advantages; this is taken up in Section \ref{methods:three-methods}.

\subsection{Idiosyncratic Clinical Language}
Specific clinical domains (e.g. neurology, psychiatry, respiratory medicine) are particularly difficult for general purpose foundation LLMs trained on biomedical research literature that tends to adhere to a more consistent and common vocabulary.  There are additional complexities and idiosyncrasies of language used in clinical practice that inevitably reflects regional- or specialty-specific nomenclature; not least, the long established differences between the UK and US in both concepts and language used in diagnostics \cite{kendellDiagnosticCriteriaAmerican1971} and the healthcare system's administrative processes can impact on e.g. the prevalence of certain conditions \cite{stringarisUnpackingDifferencesUS2014}.  As the name suggests, multidisciplinary teams (MDTs) are composed of different professionals and in mental health, a patient may be assessed or examined by different clinicians (doctors, nurses, social workers, psychologists). Each professional is educated and trained to focus on and deliver different aspects of patient care and this will be reflected in the language recorded in the patient's EHR -- for example, in one systematic review, doctors clinical reasoning content was found to be more narrowly focused on establishing a diagnosis and management plan leveraging theoretical knowledge (with the patient being the object of reasoning), whereas nurses focused on understanding the current needs of the patient, often in the context of their relatives and community \cite{vreugdenhilReasoningDoctorNurse2023}.  Even in well-constrained use cases, such as pharmaco-vigilance for medication adverse events, clinical language patterns, idioms and idiosyncrasies in EHR data are notoriously difficult to work with \cite{luo_national_2020}.

Similarly, consultation with clinical colleagues in the United Kingdom's NHS suggests that the routine clinical language recorded in EHR systems differs substantially from that used in large open-source text datasets of which most, if not all, popular open-source LLMs are trained. Whilst numerous biomedical or clinically trained LLMs exist, there is no publicly available LLM for the specific UK-based ``NHS mental health language'' and most use data from the United States.

There are several techniques to mitigate these problems which typically rely upon a form of transfer learning or domain adaption. One typical approach is to continue training the LLM in this specialist domain using the same language modelling objective, to better prepare the model for deployment in the new domain, which has delivered promising results for US based clinical datasets \cite{huang2019clinicalbert, alsentzer-etal-2019-publicly, taylor_clinical_2023}.

\subsection{Redundancy within Clinical Text}
EHR clinical text is known to be large in volume with a great deal of redundancy, repeated information, and clinically irrelevant information \cite{tayefi_challenges_2021, 2015-clinical-nlp-challenging, searle_estimating_2021}. We endeavour to implement approaches that can automatically select or \textit{attend} to the most relevant clinical information without any annotation required. Thus, we hope to show that one can design a system to ingest all unaltered clinical text and extract the meaningful signal to aid downstream clinical applications.

\subsection{Efficiency with LLMs}
A crucial element of any pipeline utilising LLMs is efficiency, especially with regards to clinical settings where compute resources can be low, training data scarce and transparency is required. The latest LLMs such as: Llama-2 from Meta \cite{touvron_llama_2023}, GPT-3 and 4 from OpenAI \cite{openai_gpt-4_2023} and PaLM \cite{chowdhery_palm_2022} use hundreds of billions of parameters. These models require significant (but largely undisclosed) financial, compute and time resource to train.  To deploy these models for inference alone can require substantial computational resource, seldom available when processing confidential clinical datasets.  Therefore, we sought to develop LLM pipelines that are tractable on limited compute budgets, and also enable further training to fine-tune on a given downstream task in an end-to-end fashion.

\subsection{Related Work}
A number of studies have utilised \textit{structured} EHR data to representation-learn patient embeddings\cite{wu_comparison_2022, rasmy_med-bert_2021} pertaining to acute or general hospital units. These studies do not, however, use or ingest unstructured free text clinical notes, instead relying on available structured fields. Several studies have utilised the MIMIC-III dataset to develop long sequence transformer-based approaches to predict different clinical outcomes and ICD-9 diagnosis codes \cite{li_clinical-longformer_2022, van_aken_clinical_2021, huang_plm-icd_2022}, but virtually none have investigated mental health-specific EHR data. 

\subsection{Desiderata for LLM Assisted Triage}
\label{sec:desiderata}
In summary, our approach to triage assistance using LLMs must address the following considerations:
\begin{enumerate}
    \item \textbf{End-to-end ingestion} of unstructured clinical EHR text to assist in triage, capable of gracefully handling variable-length inputs (e.g. at the document, referral and instance level) while maintaining classification performance
    \item \textbf{Resource efficient} in GPU compute and memory requirements such that re-purposing of foundation LLMs is feasible for the specific clinical use-case
    \item \textbf{Ability to interrogate models} to present users of assisted triage with evidence from the source EHR data that drives a triage recommendation -- in accordance with guidance on people's right to explanation for a decision making use of AI assistance \cite{ExplainingDecisionsMade2024}
    \item \textbf{Facility to train at-scale} without the need for human expert annotation of e.g. entities, concepts or text thought to be relevant to a triage decision 
\end{enumerate}

\section{Methods}
\label{sec:methods}

\subsection{Dataset}
\label{sec:methods-dataset}

For this study, we used electronic health record data from patients in Oxford Health NHS Foundation Trust (OHFT), a regional UK-based provider of specialist secondary mental healthcare to Oxfordshire and Buckinghamshire's population of around 1.2 million people.  From OHFT's EHR, we have access to historical data for approximately 200,000 patients spanning over a decade, with a total of around 8 million de-identified, pseudonymised clinical notes.  

Alongside the narrative clinical notes, we can access structured information related to referral date, which team accepted or rejected a referral (although this data is not always reliable), the dates patients were discharged after an episode of care and certain demographic information.  The routine use of these structured EHR fields is, however, subject to variability in practices between different teams;  so, rather than rely on these structured fields, we combine structured information (related to a patients referral date) and subsequent discharge date to establish whether a team accepted (or did not accept) the patient. We developed a heuristic rule in collaboration with clinicians at OHFT to remove dependency on an often unreliable field used to record if a patient was ``accepted'', ``rejected'' or when the use of ``discharge'' date is used as a proxy for a rejected referral. The  heuristic for an accepted referral is as follows: for any given referral we extract the EHR structured \textit{referral date} and determine whether the referral was closed or left open after a 14 day cut-off.  We treat every referral instance independently, so a patient may be referred to another team (i.e. rejected from one team, but referred on to another), and these would be seen as two separate referral instances. A referral instance that was within 14 days of the data extraction date (i.e. right-censored patients) was removed due to the inability to determine their acceptance to the referred team.  We adopted this heuristic because referral dates are well-recorded and it is reasonable to assume in the NHS that if no notes are entered in the window 14 days after a referral that the patient has not been accepted.  Applying this heuristic, we derived training and evaluation samples over 5 sub-specialty teams, detailed in Table \ref{tab:dataset-details}.

\begin{figure}[htp]
    \centering
    \includegraphics[scale=0.6,trim={0 0cm 0 0}]{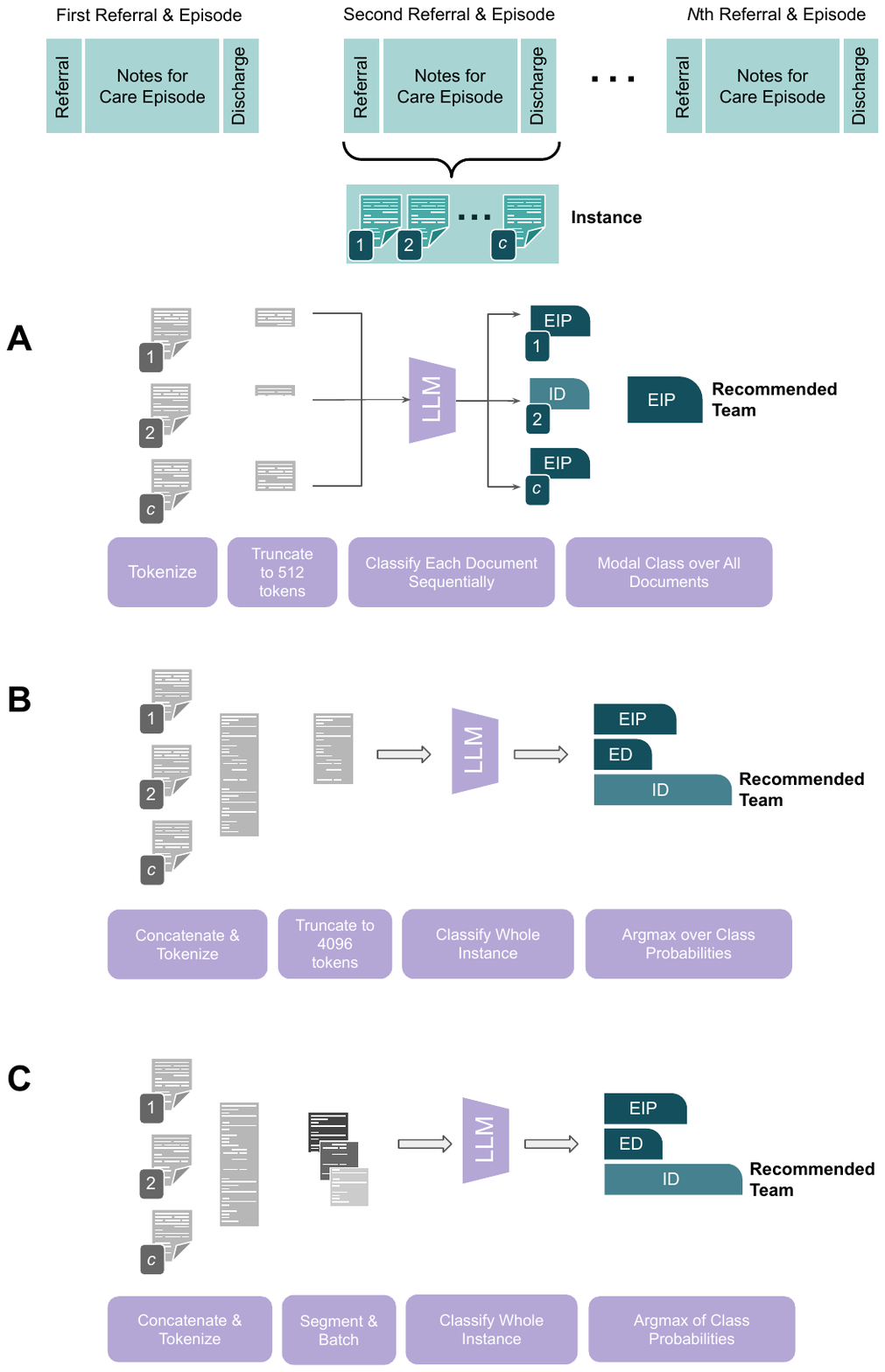}
    \caption{See main text for description of each approach}
    \label{fig:three-methods}
\end{figure}

For any given patient, we are then able to segment their entire historical EHR data capturing a collection of narrative, clinical notes demarcated by a referral and discharge date (see Figure \ref{fig:three-methods}, top panel).  We describe a collection of notes over such an episode -- consisting of a referral, zero or more documents describing the consequent episode of care up-to the discharge date -- as an \textbf{instance}. A patient may be represented as a number of instances in the training or validation samples, but not both.  Importantly, an instance consists of a time-ordered sequence of a variable number of documents (notes), each of variable word-length.  For patients who are ``routed'' (or ``bounced'') between teams, they are represented (in our data) as separate instances.   Consequently, for any individual patient, they may be represented as a number of instances but the patients (and their instances) will only be present in either the training or validation data sets.

We developed a classification task for identifying the \textit{accepted} triage team for any given instance based on the available clinical notes alone. In essence, for training, we subset only the referral instances that resulted in an \textit{accept} decision. This downstream classification task assists representation learning of instance embeddings (in the LLM) that represent referral acceptance.

\paragraph{Sub-specialty Teams}
\label{sec:sub-specialty-teams}
Our initial approach adopted the ``single point of access'' (SPoA) triage process shown in Figure \ref{fig:triage-description} -- however, this presents both a technical and ethical difficulty.  From a technical perspective, in our current EHR data sample (from OHFT) community mental health teams (CMHTs) can act as a SPoA for triaging any patient referred into secondary care (via that team because of e.g. it's geographical coverage) \textit{as well as} being a secondary care team that accepts any patients that are \textit{not} deemed more suitable for onward referral to a number of sub-specialist community teams (detailed below).  Further, it was not uncommon for referrals to be direct to sub-specialty teams (i.e. bypassing a SPoA model) which led to validation confusion matrices showing these `general' CMHTs arbitrarily accepting patients that were known to be accepted by sub-specialty teams, see appendix \ref{appendix:referral-bounce}.  In essence, ``referral bouncing'' can be a legitimate clinical decision, or the unfortunate practice of a number of different teams declining to accept a patient for reasons which often remain opaque but are frequently related to clinicians' disagreement about which is the ``most suitable'' team.  

The ethical concern is that if we were to train a model to act as a SPoA triage assistant, then we would need to additionally include a classification for ``rejection'' from any/all teams and this risks the model maladaptively learning to `default to rejection', rather than learning the ``signal'' that drives being accepted by a team; this risks recapitulating existing critiques of the triage system in mental health \cite{chew-graham_qualitative_2007} and defeats the desirable property of being able to interrogate the triage system to understand why a particular patient has been recommended for acceptance by a specific team.  

To this end, we present our results on triaging assistance for well-defined sub-specialty teams present in adult mental health services across the United Kingdom, namely: eating disorders, mental health for people with intellectual disability, older-adults (including memory and dementia services), early intervention for psychosis and peri-natal psychiatry.  Teams that we did not include were crisis  resolution and home-treatment teams (because the referral processes and criteria are very different, representing the urgency and acuity of these patients), teams that do not provide community-based assessment and treatment (for example, psychological medicine or general hospital liaison psychiatry), and specialist research-led clinics (that will not be available in a majority of NHS trusts). The number of referral teams in our dataset was highly imbalanced; for example, teams serving older adults (usually, those over the age of 65) were over-represented and this reflects the organisation of clinical services from which the data was extracted.   Given our stated intention to develop an end-to-end triage assistance tool, we decided not to artificially balance the data set (for either training or testing).









\subsection{Ethical Approval}
The data sourced from Oxford Health NHS Foundation Trust (OHFT) are de-identified and were obtained through the Clinical Record Interactive Search (CRIS) system powered by AkriviaHealth that provides a secure data environment and platform for processing de-identified clinical case notes.  Access to and use of these de-identified patient records from the CRIS platform has been granted exemption by the NHS Health Research Authority (March 2020) for research reuse of routinely collected clinical data.  The project was reviewed and approved by the Oversight Committee of the Oxford Health NHS Foundation Trust and the Research and Development Team in November 2022.

\subsection{LLM preliminaries}
\label{methods-LLMs-prelim}
All methods utilise an LLM encoder to produce representations of clinical notes contained in a single instance. LLM encoders take a sequence (length $n$) of word tokens $w \in \mathbb{Z}^{n}$ and produces a sequence of hidden representations $H \in \mathbb{R}^{d_h \times n}$ where $d_{h}$ is the dimension of the hidden representations. In other words, LLMs will ingest clinical notes and produce representations or features of the contents. The choice of LLM is important, and we opt to utilise the following models, some of which originate from our previous work\cite{taylor_etal_embeddings_2024}. Note that the RoBERTa-base-OHFT model has been domain pre-trained using a masked language modelling objective on a separate sample of OHFT EHR clinical notes. We will use this model for the main experiments and results as proved to be superior to the standard RoBERTa-base model.

\begin{enumerate}
    \item RoBERTa-base \cite{liu_roberta_2019} is a general domain LLM
    \item RoBERTa-base-OHFT is a LLM initialised from RoBERTa-base and continually pre-trained on OHFT EHR text using a masked language modelling objective
    \item Clinical Longformer \cite{li_clinical-longformer_2022} is a LLM model trained on clinical text
\end{enumerate}

To adapt these LLMs to a sequence classification task, which in essence is how we will treat our triage sub-team task, we utilise a typical approach of using a classification head on top of the LLMs outputs\cite{taylor_clinical_2023}. In our use case of sequence or document classification, the downstream task head is a single or multi layer perceptron, often refered to as the classication head: $f_\text{head}(\cdot)$ which takes the sequence embedding output by the LLM, $e$, as input and generates an $n\text{-dimensional}$ vector, where $n$ is the number of classes. The exact algorithm for deriving the LLMs sequence output is dependent on the methods outlined below\ref{methods:three-methods}.

\subsection{Patient Referral Representation Learning - Individual documents vs Long sequence}
\label{methods:three-methods}

As mentioned, we treat the triage problem as a text sequence processing task, akin to representation learning over the patient's notes encapsulated as instances as defined previously. For patients with longer histories of interacting with secondary mental health services, their instances can amount to a large body of clinical notes: in our dataset, the number of tokens (words) ranged between 300 and 50,000; see Appendix \ref{appendix:instance_tokens} for details.  The clinical `signal' contained in each note will be highly variable, with a mixture of note categories produced by different members of the clinical team with widely varying purposes. 

To mimic the clinicians perspective of viewing these historical clinical notes, we feed each of our models the notes in the reverse chronological order, whereby most recent notes are at the beginning. The reasons are two-fold: clinicians will typically be presented the most recent documents in EHR user interfaces, and our models typically operate in a bi-directional manner and in certain scenarios will truncate very long sequences.

To cope with variable-length EHR histories, in this paper, we compare three different approaches to handling variable token sequence lengths (Figure \ref{fig:three-methods}).  Denote the $n$th instance as the time-stamp ordered set of $c$ documents: $I_n = (d_1, d_2, \cdots, d_c)$ and the output triage recommendations as a multi-class probability vector $ \Pr( \mathbf{y} | \bullet ) = (y_1, y_2, \cdots, y_{T})$ where $T$ is the number of teams considered.  We compare three different approaches to ingest and process any instance:  
\begin{enumerate}[label=(\Alph*)]
    \item a document-level \textbf{`brute force' approach}: Each document $d_i \in I_n$ is tokenised, truncated to length 512 tokens ($d^{\star}_{i}$) and passed to the LLM in order.  For each document, the LLM with the classification head $f_\text{head}$ delivers $\Pr( \mathbf{y} | d^{\star}_{i} )$ and the ``recommended'' triage team $j \in T$ for that document is a vote $v_{i} = \arg \max_{j \in T} \Pr( y_{j} | d^{\star}_{i} )$.  After ingesting an entire instance $I_n$, we have an ordered set of votes for each document $V_{I_n} = (v_1, v_2, \ldots, v_c)$ and the recommended team is the modal value over $V_{I_n}$.
    \item an instance-level \textbf{single concatenated sequence approach}: The documents $d_1, d_2, \ldots d_c$ in the instance are concatenated (in order) and the resulting instance is tokenised before being truncated at the respective models maximum token length. The resulting truncated instance $I^{\star}_{n}$ is fed through the LLM and its output passed to the classification head $f_\text{head}$ and the recommended team is $\arg \max_{j \in T} \Pr( y_{j} | I^{\star}_{n} )$.  The original transformer architecture \cite{attention-all-you-need} was limited to 512 tokens and longformer models \cite{longformer2020} advance on this to provide a maximum input sequence of 4096 tokens.  We use a standard RoBERTa-based model with 512 tokens and compare performance to a 4096-token longformer.  We did not train our own longformer model (on NHS clinical notes) and instead utilise the Clinical-Longformer model \cite{li_clinical-longformer_2022, johnson_mimic-iii_2016}.
    \item an instance-level \textbf{segment-and-batch approach}: as for (B) we concatenate the documents for an instance $d_1, d_2, \ldots d_c$ in order, tokenise, but we do not truncate the resulting sequence.  Instead, following \cite{huang_plm-icd_2022, vu_label_2020, edin_automated_2023}, we divide the tokenised instance $I^{\star}_{n}$ into an ordered sequence of non-overlapping \textit{segments} of fixed size, $s$ (here, we trialled 128, 256 and 512 tokens) where the last segment is padded as required.  The resulting sequence of segments is then treated as a \textit{batch} for feed-forward processing by the LLM to produce an output embedding for each token, which are re-organised into sequence order before being passed to the classification head $f_\text{head}$ and the recommended team for that instances is decided as for (B).  We also utilise the label-aware attentions to derive representations capable of supporting interpretable classification decisions \cite{mullenbach_explainable_2018}, detailed in \ref{appendix:explainable-decisions}.   
\end{enumerate}

Finally, to improve training efficiency (in terms of the number of trainable parameters required for adapting the LLMs to the triage task), we also use the Low-Rank-Adaptation (LoRA)\cite{hu_lora_2021} method. LoRA effectively assumes the full weight updates generated during fine-tuning are intrinsically low-rank and can be approximated via singular value decomposition. The result is the training of considerably fewer model parameters during fine-tuning, and production of adapter weights that can be easily added or removed to the initial LLM (further details about LoRA, see appendix \ref{appendix:lora-method}).  The three main approaches offer different levels of granularity and efficiency which have been summarised in Table \ref{tab:triage-method-summary}.

\begin{table*}
    \centering
    \scriptsize
    \begin{tabular}{llcccc}
        \textbf{Approach} & \textbf{Base Model}  &\textbf{ \# Params}  &\textbf{ Max length}  & \textbf{Infer. speed (SD)} \\
        \toprule
         Brute Force [\textbf{A}] &  Roberta-base & 125 mil. & 512   & 683 (0.41) \\
         Concat truncated [\textbf{B}] & Roberta-base  & 125 mil. & 512  & 14 (0.01)\\
         Concat Longformer [\textbf{B}]  & Clinical-Longformer & 148 mil.  &  4096  & 212 (0.2)\\
         Segment-batch/LoRA [\textbf{C}] & Roberta-base & 125 mil./0.8 mil.  & max*  & 97 (0.69)\\
        \bottomrule
    \end{tabular}
    \caption{Overview of different sequence representation approaches and model setups with the number of trainable parameters, maximum sequence length and inference speed. The square brackets next to each approach refers the methods outlined in Figure \ref{fig:three-methods}.  We present the inference speed averaged of 500 instances from the evaluation data reported in seconds.  The SD (standard deviation) represents variance over three repetitions. *The max length for the segment-batch approach is hardware dependent.}
    \label{tab:triage-method-summary}
\end{table*}

\section{Implementation details}\label{implementation}
\subsection{Pre-processing and Cleaning}
For language modelling with transformer-based models, minimal data cleaning is required as the tokenization of inputs paired with the \textit{contextualised} representations of words actually means we want to keep as much of the original input as possible. The pre-processing steps taken included removal of carriage returns, tabs, extra white spaces and any poorly encoded characters. We intentionally took no steps to disambiguate acronyms, or remove jargon as we felt it was best to encourage the LLMs to learn the noise commonly found in clinical text.

\subsection{Training and hardware overview}
\label{sec:implementation:hardware}
All training, inference and evaluation were carried out on a single {NVIDIA} Tesla T4 16 Gb GPU co-located with the OHFT data on a virtual machine hosted on a private Amazon Web Services (AWS) instance (emulating a minimal-resource environment).  Training and evaluation data were split on unique patient identifiers to ensure no data leakage, and details of sample numbers are provided in table \ref{tab:dataset-details}. The different modelling approaches (Table \ref{tab:triage-method-summary}): \textit{Brute force, Concat truncated, Concat Longformer and Segment-batch} each have varying compute requirements and it was not possible to align all hyperparameters during training.

\begin{table*}
    \centering
    \footnotesize
    \begin{tabular}{cccc}
    \toprule
        \textbf{Dataset}  & \textbf{\# labels} & \textbf{\# train samples} & \textbf{\# eval samples} \\
        \midrule
        Accepted Triage Team Brute.  & 5 & 65,000 & 157, 880 \\
        Accepted Triage Team Concat. &  5 & 17,629 & 4,272 \\
         \bottomrule
    \end{tabular}
    \caption{Dataset details. Note that both datasets relate to the same full evaluation \textit{referral instances}. With the brute force approach we treat every single document separately, hence the significantly higher individual samples. Both in fact relate to the same number of referral instances. The training sample for the brute force approach was randomly sub-sampled to 13,000 samples per class.}
    \label{tab:dataset-details}
\end{table*}

\begin{table*}[h]
\centering
    \footnotesize
    \begin{tabular}{l l l l l l}
    \addlinespace
    \textbf{Parameter} &\textbf{ Brute force} & \textbf{Concat Trunc.} & \textbf{Concat Long.} & \textbf{Segment-batch} \\ \toprule
    Batch size & 8 & 8 & 1 & 1 \\ 
    Gradient accumulation steps & 2 & 2 & 16 & 16  \\
    Embedding dimension & 768 & 768 & 768  & 768 \\
    Learning rate & $1\times10^{-5}$ & $1\times10^{-5}$ & $1\times10^{-5}$& $1\times10^{-4}$ \\
    Optimiser & AdamW & Adam W & Adam W & Adam W \\
    Chunk size & - & - & - & [128, 256, 512]  \\   
    \bottomrule
    \end{tabular}
    \caption{Overview of hyperparameters used in experiments for each instance modelling approach. All training regimes utilised a linear scheduler with warm-up and early stopping with F1 score as the criteria and a patience of 3}
    \label{table:hyperparas}
\end{table*}

\section{Results}




Table \ref{tab:main-classification-results} compares the three sequence representation methods (Fig. \ref{fig:three-methods}) when utilising a) RoBERTa fine tuned on the OHFT data or the Clinical-Longformer or b) RoBERTa-base, without fine tuning on the OHFT data.  As expected, we find that overall, a model fine-tuned on the actual OHFT EHR data generally provides a marginal performance benefit irrespective of the sequence representation method used.  Of the three sequence representation methods, `segment-and-batch' (method C in Fig.\ref{fig:three-methods}) is consistently the best method.  We note that using LoRA to improve training efficiency (i.e. LoRA requires updating $< 1\%$ of base LLM's model parameters compared to fully fine-tuning the base LLM) incurs only a small degradation in F1 performance of 0.014 (e.g. when LoRA was used with the segment-and-batch approach).  Given the superior performance of the segment-and-batch approach and some benefit to using the RoBERTa-OHFT fine-tuned LLM, the approach we take forward for further analyses considers the RoBERTa-OHFT with segment-and-batch.

\begin{table}[htp]
\centering
\begin{subtable}{0.9\textwidth}
    \centering
    \footnotesize
    \label{tab:cls-roberta-ohft}
    
    \begin{tabular}{lllcccc}
    \toprule
    \textbf{Model} & \textbf{Approach }& \textbf{Accuracy} & \textbf{F1} & \textbf{Precision} & \textbf{Recall} \\
    \midrule
     RoBERTa-OHFT & Brute force [\textbf{A}]  &0.935 & 0.846 & 0.828 & 0.882 \\
    RoBERTa-OHFT & Concat trunc. [\textbf{B}] &0.974 & 0.922 &      0.927 &   0.917 \\
    Clinical-Longformer & Concat Longformer [\textbf{B}] &0.975 & 0.924 & 0.932 &   0.918 \\
    RoBERTa-OHFT &\textbf{Segment-batch} [\textbf{C}] &\textbf{ 0.981} & \textbf{0.938} &     \textbf{ 0.942} & \textbf{0.933} \\
    RoBERTa-OHFT &Segment-batch-LoRA* [\textbf{C}]& 0.968 & 0.924 & 0.927 & 0.919 \\
    \bottomrule
    \end{tabular}
    \caption{RoBERTa-OHFT}
    \end{subtable}
    \hfill
\begin{subtable}{0.9\textwidth}
    
    \centering
    \footnotesize
    \label{tab:cls-roberta-base}
    \begin{tabular}{lllcccc}
    \toprule
    \textbf{Model} & \textbf{Approach }& \textbf{Accuracy} & \textbf{F1} & \textbf{Precision} & \textbf{Recall} \\
    \midrule
    RoBERTa-base &Brute force [\textbf{A}] & 0.923 & 0.80 & 0.78 & 0.859 \\
    RoBERTa-base& Concat trunc. [\textbf{B}] & 0.889 & 0.772& 0.71 & 0.866    \\
    RoBERTa-base& \textbf{Segment-batch} [\textbf{C}]& \textbf{0.976} & \textbf{0.922} & \textbf{0.934} & \textbf{0.911} \\
    \bottomrule
    
    \end{tabular}
    \caption{RoBERTa-base}
    \end{subtable}

\caption{Accepted triage team classification metrics for each sequence representation approach. The square brackets next to the approach refers to the methods outlined in Fig \ref{fig:three-methods}.*Is the segment-batch approach with LoRA to highlight the small drop in performance compared to the standard segment-batch. }
\label{tab:main-classification-results}
\end{table}

\subsection{Effect of Sequence Length on Performance}
We introduced three methods for handling the problem of variable document and instance lengths in EHR data (Figure \ref{fig:three-methods}).  If we hypothesise that longer token sequences contain more information that is useful for classification (triaging), we might expect classification performance to vary as a function of instance length.  To test this, we examined classification performance by stratifying the validation data into short ($<128$ tokens), medium ($>128$ and $<=512$ tokens), long ($>512$ and $<=4096$ tokens), and extra long ($>4096$ tokens) sequences.   

\begin{figure}
    \centering
    \includegraphics[scale=0.7,trim={0cm 0.5cm 0 0.5cm}]{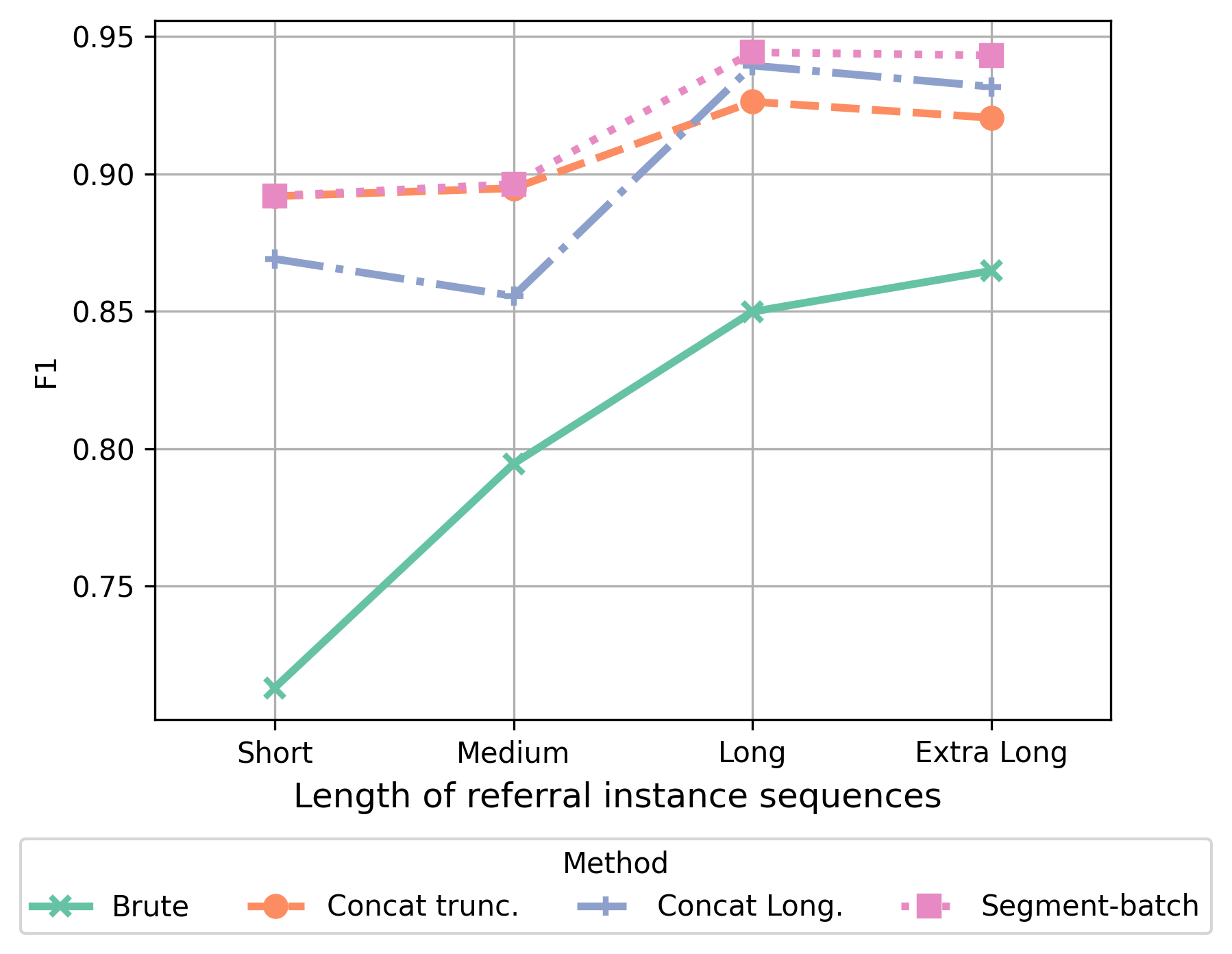}
    \caption{Validation set F1 score for each method of handling variable sequence lengths as a function of instance token lengths: short ($<128$ tokens), medium ($>128$ and $<=512$ tokens), long ($>512$ and $<=4096$ tokens), and extra long ($>4096$ tokens). 
    For details of each method, refer back to Fig \ref{fig:three-methods} where \textit{Brute} refers to method \textbf{A}, \textit{Concat trunc.} and \textit{Concat Long.} refer to \textbf{B} using standard RoBERTa architectures and the longformer respectively and \textit{Segment-batch} to method \textbf{C}.}
    \label{fig:mixed-sequence-length}
\end{figure}

In our dataset, instances resulting in acceptance or non-acceptance had a median length of 1367 and 1463 respectively.  Across all individual documents, we find that the median length is 120 tokens (IQR=155) and when considering instances (concatenations of documents), the median token length is 1323 (IQR: 3229) -- see \ref{app:triage-data-stats}. Figure \ref{fig:mixed-sequence-length} reveals that for all three methods, the longer the instance (in tokens) the better the classifier's F1 performance is.  Of note, the segment-and-batch approach is consistently as good, or better, than the other methods over all instance sequence lengths.

\section{Discussion}
\subsection{General}
In this paper, we have shown how to develop a bespoke and resource-efficient LLM for a specific healthcare task.  We described the construction of LLMs to assist triaging of patient referrals to secondary care mental health teams by end-to-end ingesting of variable-length EHR data.  We found the the segment-and-batch approach displayed the highest performance while providing the desired properties of an end-to-end system (see Section \ref{sec:desirable}).  While general purpose MLaaS LLMs have demonstrated remarkable performance on a variety of tasks, including healthcare, there remains information governance concerns about their deployment in healthcare systems.  Further, while training and deploying LLMs with the capabilities of ChatGPT, Claude and Gemini (in particular, using prompt engineering) may in the future be plausible, currently, the resource costs are prohibitive.  This means specialising and domain adapting high-performing LLMs to clinical tasks of direct relevance to healthcare remains difficult.    We propose that our approach emulates the processes that clinical teams use in community out-patient settings in the NHS i.e. when a patient is referred, referral details are documented on the EHR system alongside any existing clinical history about the patient and this EHR data is then used as the basis for a triage decision (i.e. to accept the referral as relevant to that clinical team and invite the patient for a clinical assessment). Given that 96\% of secondary specialist mental heath care is conducted in community out-patient settings, delays (and arguably, erroneous triage decisions) introduced by the referral process are a recognised problem in NHS services \cite{chew2007qualitative,HiddenWaitsForce}. 

\subsection{Recommendations}
Our results show that the most promising method for ingesting variable-length EHR data (representing patient histories) in an end-to-end fashion for a downstream task is to use the segment-and-batch approach \cite{huang_plm-icd_2022, vu_label_2020, edin_automated_2023,mullenbach_explainable_2018}.  This method enables consistent performance over a broad range of token sequence lengths as well as gracefully handling relatively short or sparse clinical documents.  We note that the other methods employed (brute-force and concatenate-truncate) all involve the compromise of truncating documents or concatenations of documents (instances) to fit the maximum input sequence length. The segment-and-batch approach is also limited to a fixed input sequence length (approximately 12,000 tokens on the GPU used, see \ref{sec:implementation:hardware}), but this is more flexible (and larger) because it is a function of the GPU's memory (i.e. constraints on the batch size) and not the architecture.  

\subsection{Evaluation against Stated Desiderata}
\label{sec:desirable}
We now describe our results with reference to the four criteria introduced in Section \ref{sec:desiderata}:
\begin{enumerate}
    \item \textbf{End-to-end ingestion}: we have shown that the segment-and-batch provides a plausible method for ingesting EHR data requiring only tokenisation, proceeding to embedding and then recommendation in a downstream classifier
    \item \textbf{Resource efficient}: using LoRA and the segment-and-batch architecture, training and inference can be managed on a single GPU (see Table \ref{tab:triage-method-summary})
    \item \textbf{Ability to interrogate models}: while a full evaluation of the interpretability of triage results delivered by our proposed model is beyond the scope of this paper, we present initial experiments showing how the segment-and-batch model can be interrogated in \ref{appendix:explainable-decisions} using what we have previously described as \textit{interpretability through presentation} \cite{joyce2023explainable}
    \item \textbf{Facility to train at-scale} with minimal human-input: our model makes use of a combination of administrative data (referral dates) and a heuristic based on the expected behaviour of clinical teams (i.e. that if there are no data entered within a time-frame after the referral date, we assume the referral was \textit{not} accepted) to deliver a training target for classifying any given instance.  Most importantly, we utilise LLMs' ability to learn a latent space of token-sequence embeddings that can be used as the basis for downstream classification to deliver a triage recommendation.  
\end{enumerate}

We note that an additional benefit of using LoRA with an LLM is that a pre-trained base LLM (here, using a RoBERTa base pre-trained on routinely-collected clinical data from an NHS clinical records system) can potentially be re-used as a general-purpose encoder for deriving embeddings for down-stream tasks for similar applications in mental healthcare -- this is because LoRA adapter weights can be held separately to the underlying base LLM, or be fully integrated through simple matrix multiplication.

\subsection{Limitations}
\paragraph{Computational resource}
Due to the nature of the dataset, we intentionally limited our experiments to consider model architectures that -- at the time of writing -- could be plausibly implemented on a modest, single GPU. Consequently, this means we were only able to utilise smaller LLMs in our experiments and we expect future work would seek to utilise more recent LLMs (e.g. Llama-2 \cite{touvron_llama_2023}). These limitations also meant that with our segment-batch approach gracefully managed upto 12,000 tokens and this limit is a function of the GPU hardware capabilities - in our case 16Gb VRAM.  

\paragraph{Alignment with Clinician Behaviours}
The derivation of which triage team a referral instance and its associated clinical notes belonged to was based on available administrative information (e.g. referral date, which is reliable in our data) and a simple heuristic for \textit{acceptance}. Whilst this was developed with clinicians with knowledge of the clinical culture and practices of the source data (OHFT), we did not obtain any gold-standard decisions for a sample of clinical notes and future work will need to compare the performance, utility and alignment of the triage assistance system with clinical practice.  

\paragraph{Privacy of trained models}
Due to the nature of the training data, we are unable to share the underlying LLM models or classification models developed for this work. Any pipelines involving LLMs and confidential training data have inherent data security issues, with LLMs increasingly showing capabilities to leak, or even re-create training instances \cite{carlini_extracting_2021, carlini_poisoning_2023}. Therefore, to scale these approaches and apply to external datasets would require careful governance to allow any model sharing. 

\subsection{Future Work and Directions}
Future work is required to:
\begin{itemize}
    \item test the acceptability and utility of the ``interpretability by presentation'' model (shown in \ref{appendix:explainable-decisions}) with clinicians trialling the triage assistance system using a mixed-methods study of clinician's behaviour when using the system 
    \item instead of a single `monolithic' multi-team triage system, there may be benefit to implementing an ensemble of triage `agents' each specialising in detecting and representing signal in instances (referrals and patients) specifically that team.  The recommended team would then be a function of the recommendations produced by the ensemble.  We note that this rehearses the ethical question described in Section \ref{sec:sub-specialty-teams} because each agent would then need to be trained to accept \textit{or reject} a referral; if data derived from a particular team's triaging behaviour is biased and inequitable (e.g. for certain kinds of patients, there is a disproportionate probability of rejection) then the ensemble's performance will begin to reflect these same patterns.  
\end{itemize}

\section*{Acknowledgements}
 We would like to acknowledge the work and support of the Oxford Research Informatics Team: Tanya Smith, Research Informatics Manager, Adam Pill, Suzanne Fisher, Research Informatics Systems Analysts and Lulu Kane Research Informatics Administrator.
\section*{Funding}
NT was supported by the EPSRC Center for Doctoral Training in Health Data Science (EP/S02428X/1). AK, ANH, IL and DWJ were supported in part by the NIHR AI Award for Health and Social Care (AI-AWARD02183). DWJ is part supported by an NIHR Infrastructure Programme (NIHR203316).  AC is supported by NIHR Oxford Cognitive Health Clinical Research Facility, by an NIHR Research Professorship (grant RP-2017-08-ST2-006), by the NIHR Oxford and Thames Valley Applied Research Collaboration, by the NIHR Oxford Health Biomedical Research Centre (grant NIHR203316) and by the Wellcome Trust (GALENOS Project). The views expressed are those of the authors and not necessarily those of the UK National Health Service, the NIHR, or the UK Department of Health.

\section*{Contributions}
N.T, D.W.J, A.K, and A.N.J conceptualised this work. N.T and D.W.J curated the datasets. N.T developed pre-processing and experiment running and analysis code. N.T and D.W.J drafted the manuscript. A.K, and A.N.H, I.L, and A.C revised and edited the manuscript. All authors read and approved the final version of the manuscript.

\bibliographystyle{unsrtnat}
\bibliography{references.bib}

\appendix




\section{Triage and Team Referral Bouncing}
\label{appendix:referral-bounce}
The most common pathway for patients to be assessed and then treated in specialist secondary care is for another professional (often, a primary care general practitioner / family physician) to send a referral to a local secondary care organisation.  Assume Team A receives a referral, but decides that another service -- Team B -- would be better positioned to assess or treat the patient.  Team A will then forward the referral to Team B.  On receipt of the referral, Team B might conclude that either i) another service, Team C, is better suited to care for the patient or ii) that Team A should have accepted the patient's referral in the first place.  This results in cycles of what is termed ``referral bouncing'' and is an unfortunate result of service pressures and -- as described by \cite{chew-graham_qualitative_2007} -- ``capricious'' and ``opaque'' decision making that in reality reflects arbitrary application of referral criteria in the interests of the team (rather than the patient).  Of course, there are legitimate clinical reasons for referral forwarding and bouncing; in some clinical services, a team might function as a single-point-of-access for a geographical region (see Figure \ref{fig:triage-description}) in addition to having an assessment/treatment function for the same cohort of patients.

\begin{figure}
    \centering
    \includegraphics[scale=0.5]{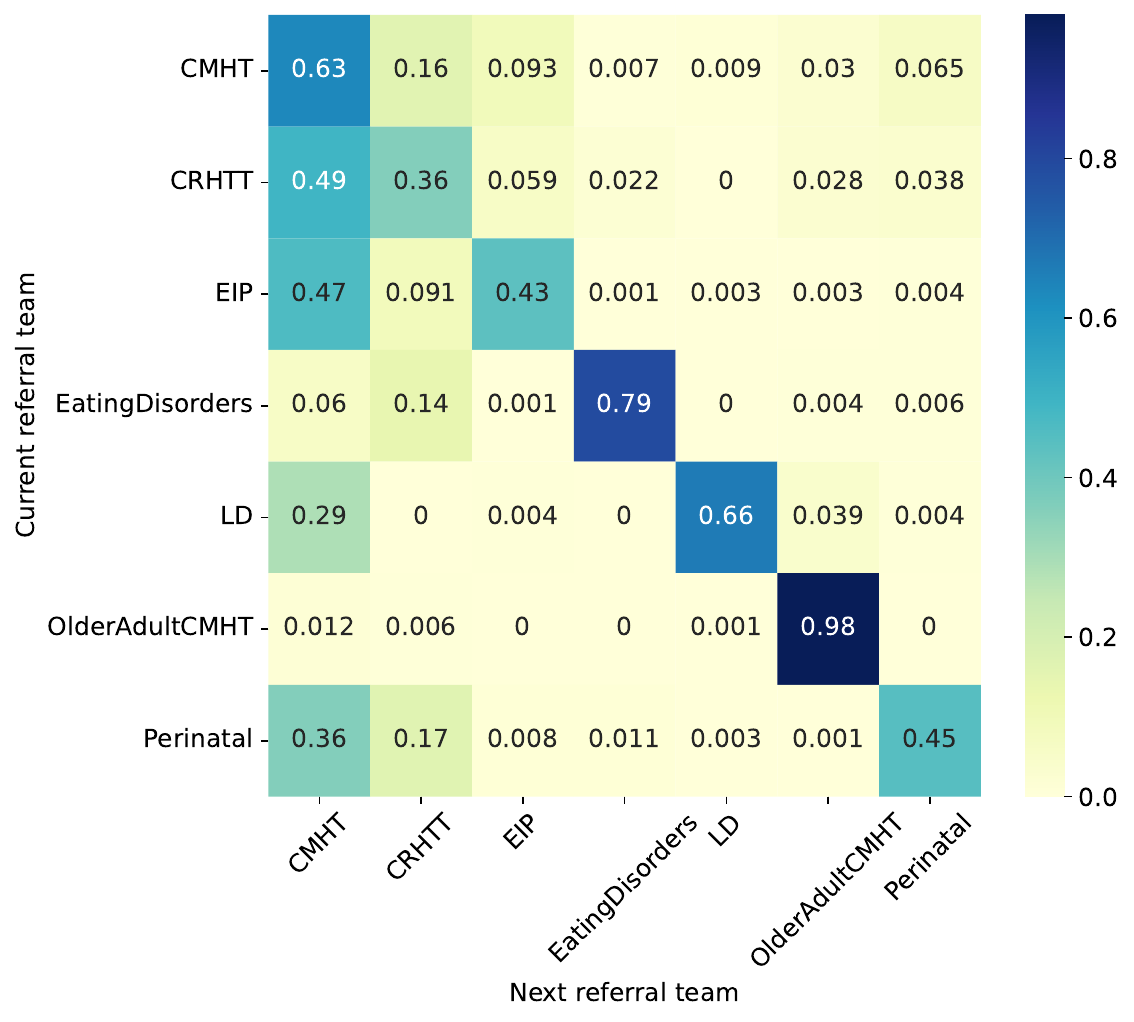}
    \caption{Tabulation of the probability of the team first receiving a referral (Team A, rows) referring the patient onto another team (team B, columns) within a 30 day window.}
    \label{fig:referral-crosstab}
\end{figure}

Liaising with clinicians working in the secondary care system from which our data originated, it was clear that some teams (notably, community mental health teams designated ``CMHTs'') have these both these functions.  Therefore, it is not uncommon for a referral to appear in the EHR as a referral to the CMHT and then quickly, as another referral to a different team.  Figure \ref{fig:referral-crosstab} shows a descriptive analysis of these initial first- and second-referral patterns in our data set.  As a concrete example; CMHTs frequently receive referrals for patients in crisis, for which they (reasonably) refer to CRHTT (crisis resolution and home treatment) teams.  CRHTTs are specifically designed to be disorder agnostic and to help patients within hours of being referred and to specifically address and manage crisis presentations that are not (in general) the remit of CMHTs or other secondary care teams.  This is reflected in the first column of Figure \ref{fig:referral-crosstab}.  Similarly, by examining the diagonal of Figure \ref{fig:referral-crosstab}, it can be seen that for some teams (notably, teams working with older adults) there is a high probability that on first being referred, that referral will remain with that team.  The asymmetry in referral patterns is equally revealing.  For example, a patient referred to a sub-specialty team for Early Intervention in Psychosis (the EIP row in \ref{fig:referral-crosstab}) has probability of 0.47 and 0.43 of being referred to a ``general'' CMHT and CRHTT respectively.  The former may suggest an inappropriate referral (i.e. the patient does not present meet the team's criteria of being a first episode of a psychotic disorder) while the latter reflects that many crisis presentations have features that would (in the absence of crisis) have been suggestive of a psychotic illness that EIP teams are specifically designed to help.  

Finally, we note that using LLMs and classification in assisting triage is fundamentally an inductive learning problem using a discriminative model -- we attempt to learn the probability of a team accepting a referral (from data contained in the EHR) given the learned patient representations (from free-text, narrative documentation in the EHR).  The data contained in the EHR does not explicitly describe the clinical reasoning that leads to an acceptance (or rejection) of a referral in a way that can be interrogated or exploited -- so at the point of referral (e.g. within a specific instance -- the fundamental unit of input to the triage assistance system -- see Figure \ref{fig:three-methods}) we cannot conclude that a referral was bounced for a clearly clinical reason (e.g. the patient was in crisis and appropriately forwarded to the crisis team), the referral was incorrectly sent to that team (i.e. the receiving CMHT was not the correct team for that patient because of their geographical location) or if the forwarding (bouncing) of a referral results from opaque clinical reasoning/decisions.

\section{Dataset Details}
\label{app:triage-data-stats}
\subsection{Justifying the Acceptance Heuristic}
\begin{figure}[htp]
    \centering
    \includegraphics[scale=0.2]{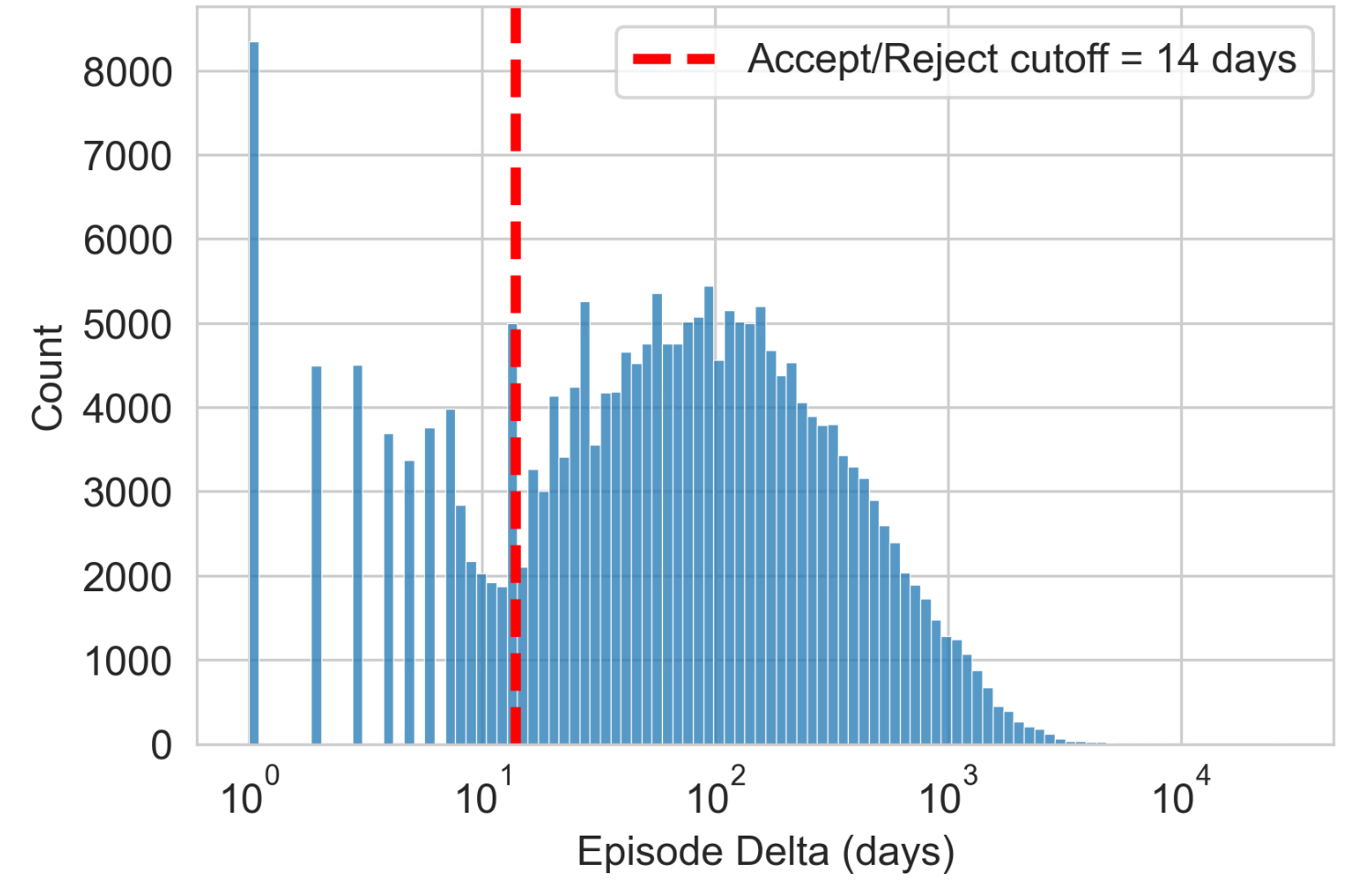}
    \caption{Histogram of the number of days individual referral instances remain open based on the available referral date, and discharge date structured fields within the OHFT EHR data.}
    \label{appendix:referral_episode_dist}
\end{figure}

As noted in Section \ref{sec:methods-dataset}, using local knowledge of the NHS and the specific clinical services available in Oxford Health NHS Foundation Trust, we were able to derive a heuristic to yield target labels for which team accepted a referral for any given instance in the training and evaluation data.  Recall that in the EHR data, structured referral date fields are generally populated and reliable, but rejection date fields are unreliable with discharge date fields often as a proxy for referral rejection.  To evidence the basis for our ``14 day rule'', we present the distribution of instance durations, based on the provided \textit{referral date} and \textit{discharge date} structured administrative data in Fig \ref{appendix:referral_episode_dist}. We see there is a large proportion of referrals that are discharged within the same day, and a slight peak around $14$ days.

\subsection{Referral Instance and Document Statistics}
\paragraph{Token numbers per instance}
Table \ref{appendix:instance_tokens} shows summary statistics for the distribution of token numbers per individual document (which limits the inputs used for the brute-force approach), and when concatenated together to form an entire instance (which limits the longformer and segment-and-batch approaches).  We note that the individual documents are generally shorter than the maximum token length for the RoBERTa-base models used in this work, whereas the full instances are substantially longer. 


\begin{table}[ht]
    \centering
    \begin{tabular}{ccc}
      Type & Mean &  Percentiles (25:50:75:90) \\
      \toprule
        Per document tokens  & 183 &  62 : 120 : 217 : 388 \\
        Concatenated instance tokens  & 6420 & 429 : 1323 : 3658 : 11427 \\
    \end{tabular}
    \caption{Descriptive statistics about the number of tokens across clinical notes related to individual referral instances.}
    \label{appendix:instance_tokens}
\end{table}



\begin{figure}[htp]
    \centering
    \includegraphics[scale=0.2]{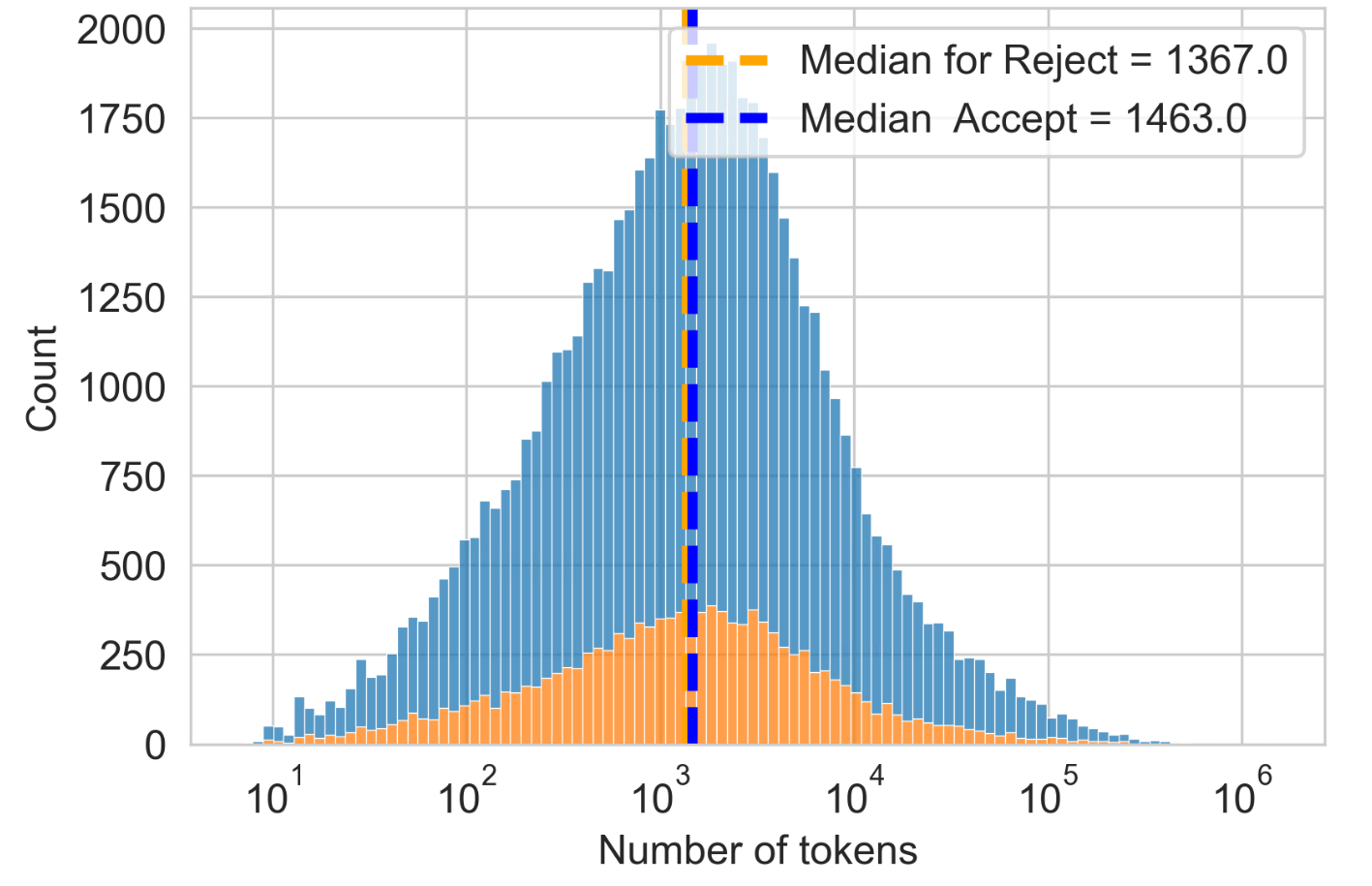}
    \caption{Histogram of the number of tokens per referral ($log10$ scale) instance based on reject or accept label}
    \label{appendix:concat_token_num_label}
\end{figure}

In Figure \ref{appendix:concat_token_num_label} we show the distribution of token numbers, per instance, as a function of whether a the referral instance was deemed to have been accepted or rejected, according to our heuristic.  We found no relevant difference between those instances that were accepted versus rejected, with a median of $1463$ and $1367$ tokens respectively. 

\section{Toward Interpretable Triage Recommendations}
\label{appendix:explainable-decisions}
We have previously argued \cite{joyce2023explainable} that it may not be possible to provide mechanistic or intrinsic interpretability for contemporary AI systems built from components that necessarily make use of `black box' methods and LLMs with downstream classification tasks are an example of such a use-case.  We have instead proposed that \textit{interpretability through presentation} may provide a way to offer clinicians the facility to interrogate decisions using such systems.  In essence, this approach tries to capture how the overall system operates (from input to output) by exposing key stages or steps in the computational process as graphical intuitions.  In our specific case, the process is a) an instance is ingested and mapped to a location in the 768-dimensional embedding space b) a mapping is learned from this embedding space to the probability of being accepted by one of 5 sub-specialty teams.  To present this process to users:
\begin{itemize}
    \item we use dimensionality reduction \cite{maaten_visualizing_2008} to display a planar projection of the 768-dimensional embedding space of the training data; this effectively provides the user with a map of the population of instances (patient referrals) emphasising the clustering of similar referral instances (and the teams they were respectively triaged to).  This gives the user the ability to visualise how similar / different the current ``query'' patient is to others known to have been accepted by the different teams. 
    \item we exploit the label-aware attention weights (of the segment-and-batch approach to ingesting and classifying instances, Fig \ref{fig:three-methods}, panel \textbf{C}) to visually highlight which instance tokens (or groups of tokens) contributed most weight to that instance's classification.  This enables the user to inspect what `source' information is driving the triage recommendation that may, for example, be useful for quickly locating data that justifies the final clinical triage decision from a multi-disciplinary team.
\end{itemize}

We propose that this gives users (clinicians) the ability to see how ``prototypical'' the patient is (with respect to the recommended team) and to locate the clinical text (feature importance) that drives the recommendation.  In what follows, We present a prototype user-interface and show how different `types' of clinical notes are handled by our assisted triage model.  It is important to note that the example clinical notes displayed here were written by an author of this paper (a clinician) to model typical examples of the kinds of notes seen in practice but they are fictional and we \textbf{do not present any confidential patient data from our data set}.

\begin{figure}[htp]
    \centering

    \includegraphics[scale=0.6,trim={0cm 4.5cm 0 0.5cm}]{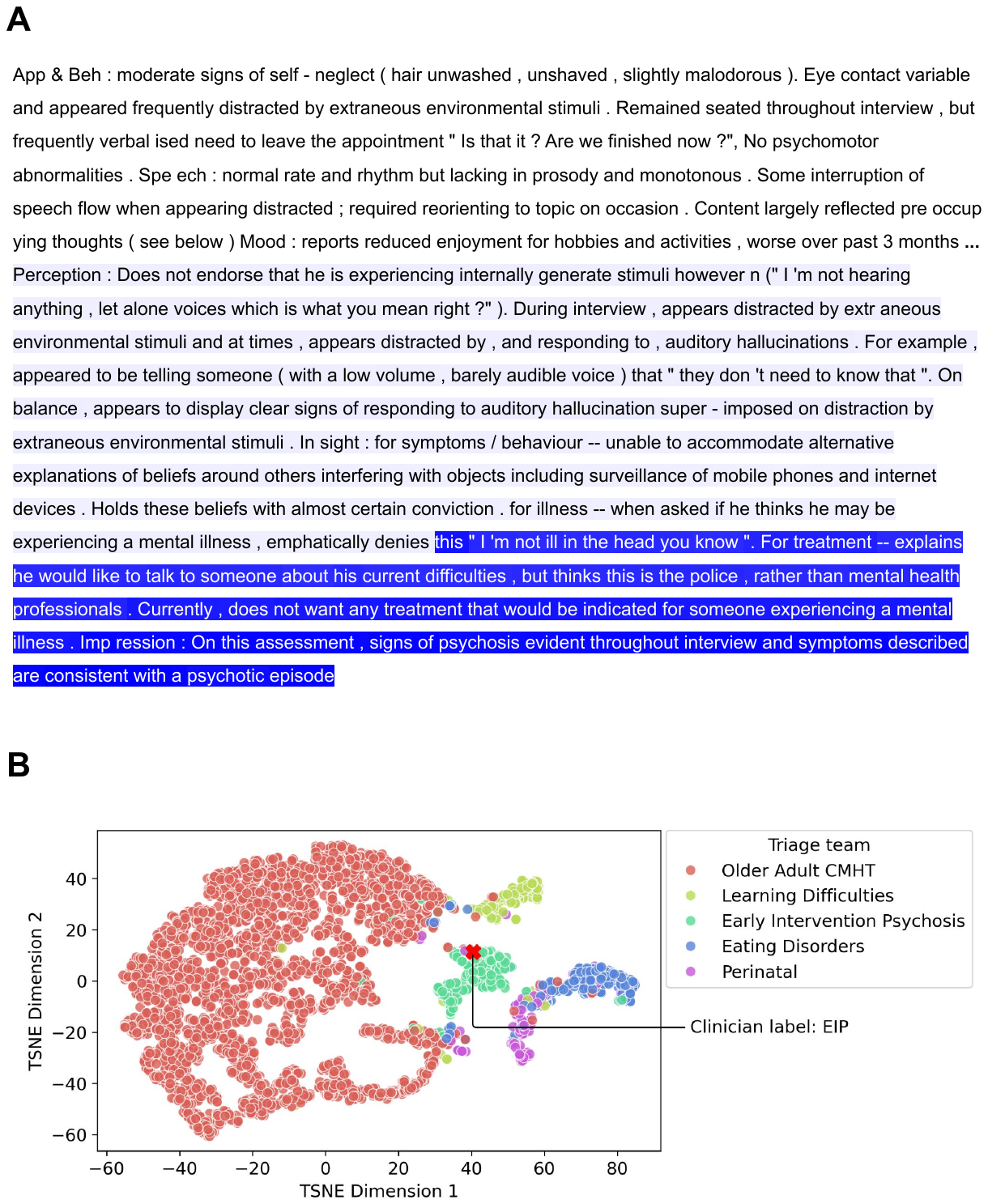}
    \caption{Mental State Exam (MSE): \textbf{A} Visualisation of label-aware attention applied to the original synthetic text, where darker blue indicates \textit{higher} soft-maxed attention scores. \textbf{B} planar projection (via t-SNE) of the training data set instance embeddings, with the query instance shown as a red-cross.}
    \label{appendix:synthetic-mse-results}
\end{figure}

\begin{figure}[htp]
    \centering

    \includegraphics[scale=0.6,trim={0cm 9.5cm 0 0.5cm}]{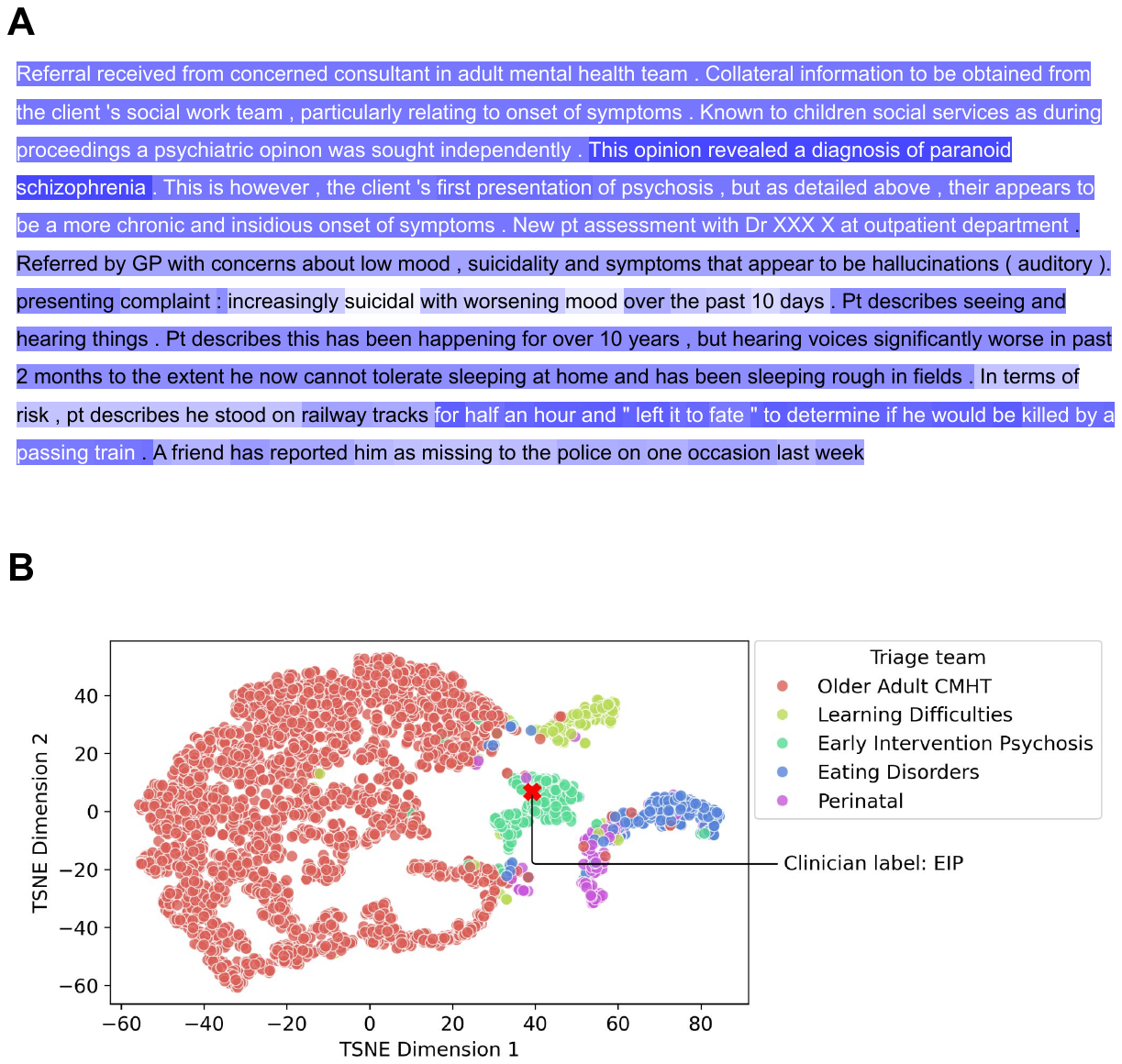}
    \caption{Summary note from an MDT meeting or discussion: \textbf{A} Visualisation of label-aware attention applied to the original synthetic text, where darker blue indicates \textit{higher} soft-maxed attention scores. \textbf{B} planar projection (via t-SNE) of the training data set instance embeddings, with the query instance shown as a red-cross.}
    \label{appendix:synthetic-eip-results}
\end{figure}

\begin{figure}[htp]
    \centering

    \includegraphics[scale=0.6,trim={0cm 2.7cm 0 0.5cm}]{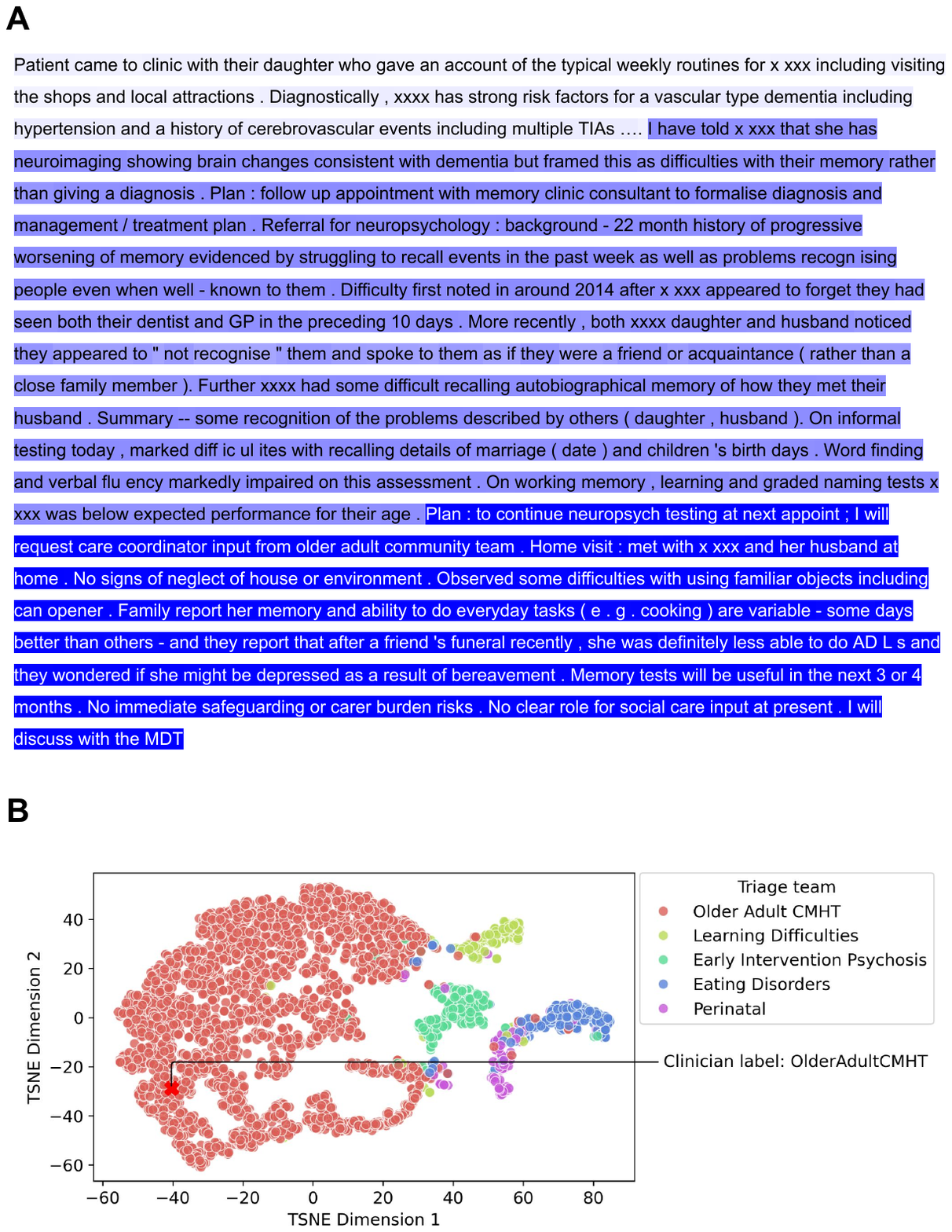}
    \caption{A short instance summarising a patient from different healthcare professionals: \textbf{A} Visualisation of label-aware attention applied to the original synthetic text, where darker blue indicates \textit{higher} soft-maxed attention scores. \textbf{B} planar projection (via t-SNE) of the training data set intance embeddings, with the query instance shown as a red-cross.}
    \label{appendix:synthetic-oacmht-results}
\end{figure}

\begin{figure}[htp]
    \centering

    \includegraphics[scale=0.6,trim={0cm 13.5cm 0 0.5cm}]{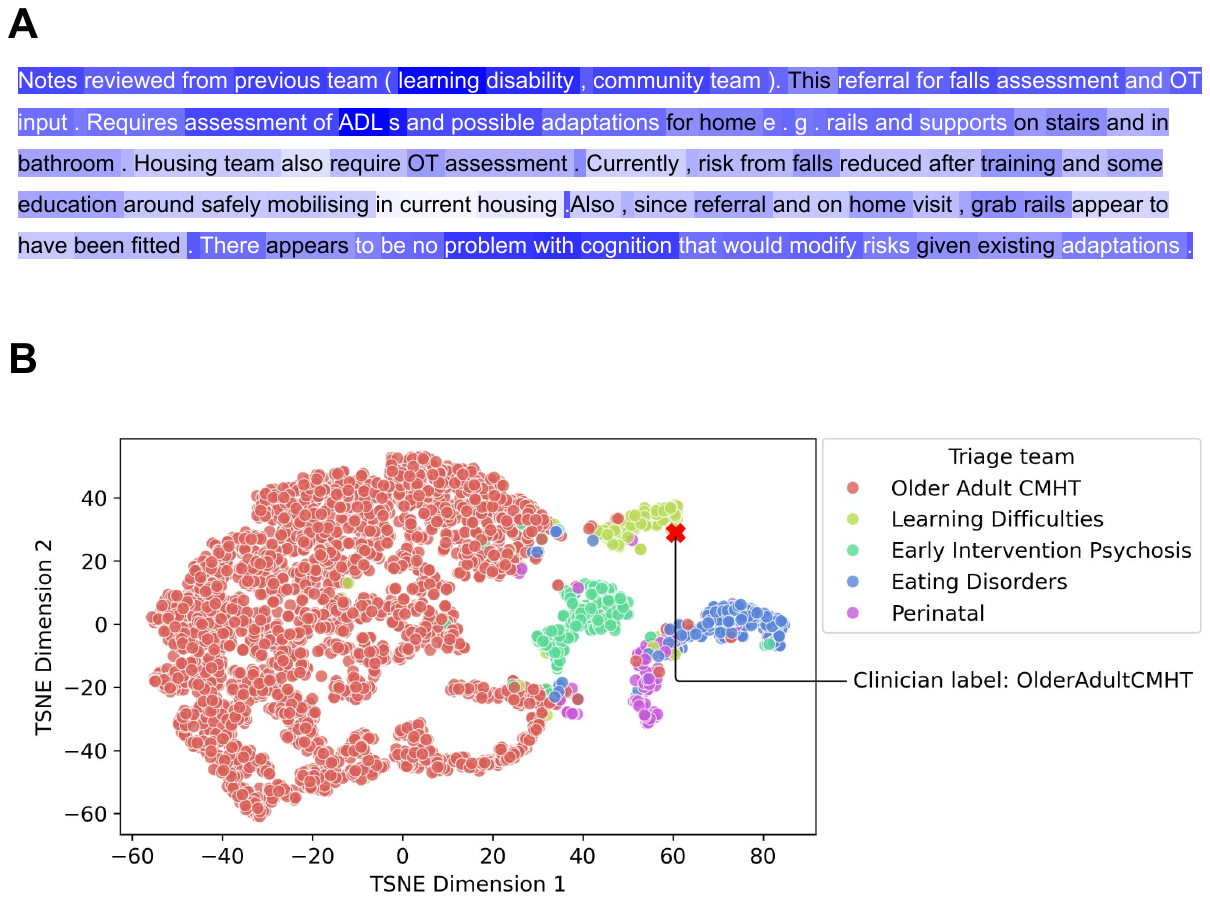}
    \caption{An administrative note highlighting previous history with one team (learning disability) but where the content reflects needs appropriate to a different (older adult) team: \textbf{A} Visualisation of label-aware attention applied to the original synthetic text, where darker blue indicates \textit{higher} soft-maxed attention scores. \textbf{B} planar projection (via t-SNE) of the training data set intance embeddings, with the query instance shown as a red-cross.}
    \label{appendix:synthetic-difficult-results}
\end{figure}

We present four examples, constructed to illustrate the following kinds of clinical notes:
\begin{enumerate}
    \item A \textbf{mental state examination} that strongly implies the patient is experiencing a psychotic episode.  A reasonable triage recommendation would therefore be an early intervention for psychosis (EIP) team.  The mental state examination is a summary of psychopathology (signs and symptoms) describing a `snapshot' or cross-section of the patient's clinical state at the time they were examined and is primarily used by doctors (e.g. psychiatrists). See Fig \ref{appendix:synthetic-mse-results}.
    \item A clinical note summarising the reason for referral, a brief statement about the patient's history and summarising the outcome of a clinical review (with salient or headline psychopathology highlighted in less formal language) written by a third-person.  This would be typical of a clinical note recording or \textbf{summarising a multi-disciplinary meeting} about the patient and may be written by a healthcare professional or administrator.  This example is presented as a likely referral to the early intervention psychosis (EIP) team. See Fig \ref{appendix:synthetic-eip-results}.
    \item An instance containing \textbf{three short summative notes from different healthcare professionals} -- 1) the first outlining a history of clinical and imaging findings and plan followed by 2) a summary of a referral for assessment by a neuropsychologist describing collateral history and summarising an initial assessment/examination and plan followed by 3) a summary of a domicillary visit with observations from e.g. an occupational therapist.  This instance would strongly imply that this patient should be recommended for an older adult team. See Fig \ref{appendix:synthetic-oacmht-results}.
    \item A brief note containing data where there is evidence of historical episodes of care with a different team (a learning disability team) but where the focus suggests a need for occupational therapy and suggests frailty that might suggest the patient should be under the care of an older adult team.  This represents an \textbf{administrative note} where superficially, one might expect the previous care team to see the patient again, as they had done previously. See Fig \ref{appendix:synthetic-difficult-results}.   
\end{enumerate}

We emphasise that the example instances shown are essentially very short instances (i.e. they represent at most three sequential clinical notes in the patient's EHR, presented without the context of a complete instance containing any historical notes) so the performance of the system shown on these test cases represents the bare minimum that can be expected.  

All results shown are from the segment-and-batch approach.  Each of the following model outputs shows:
\begin{itemize}
    \item Panel (A) displays the clinical note, with highlighting proportional to the model's label-aware attention weights.  Dark blue highlighting indicates strings of tokens that are salient and driving the classification (team recommendation) whereas lighter blue strings are considered less important.
    \item Panel (B) displays the planar projection of the embeddings of all patients' instances over the training data, with a red cross indicating the relative location of the query instance (Panel A).
\end{itemize}

These examples highlight that the model can seemingly highlight salient pieces of information relating to a triage team classification decision. We included examples which were deemed clear team-specific examples: Early Intervention Psychosis (EIP) in Fig \ref{appendix:synthetic-eip-results}, where greatest \textit{attention} appear related to clear signs of schizophrenia, and older adult community mental health team (oaCMHT) in Fig \ref{appendix:synthetic-oacmht-results}, where attention is given to a large portion of the note with an emphasis on the plan related to memory problems and next steps.



\section{LoRA}
\label{appendix:lora-method}
LoRA (Low-Rank Adaptation) of LLMs \cite{hu_lora_2021} is a reparameterization technique that works by injecting two trainable matrices ($A$ and $B$) that act as an approximation of a singular value decomposition (SVD) of the weight update $\Delta W$ for any weight matrix $W \in \mathbb{R}^{d \times k}$ in the LLM. The approximation works as $\Delta W \approx BA$, where $B \in \mathbb{R}^{d \times r}$, $A \in \mathbb{R}^{r \times k}$ and $r << \text{min}(d, k)$ is the rank of the LoRA matrices, which is a tunable parameter.

The new forward pass is updated to $h = (W + \Delta W)x \approx (W + AB)x = Wx + ABx$. While LoRA can be introduced in any layer of the LLM, it is common to use it to approximate the key, query and value matrices in the transformer architecture. This is based on the assumption that weight updates in LLMs have an intrinsically low rank compared to their dimensions, and can thus be well-approximated by their SVD.

Additionally, once trained, the LoRA matrices $A$ and $B$ can be integrated into the model as $W_{\text{updated}} = W_0 + BA$, thereby introducing no inference latency. As with other efficient training methods, the original weight matrices $W$ of the LLM remain frozen.


\end{document}